\documentclass{article}

\usepackage[english]{babel}
\usepackage{microtype}

\usepackage[letterpaper,top=2cm,bottom=2cm,left=3cm,right=3cm,marginparwidth=1.75cm]{geometry}

\usepackage{amsmath}
\usepackage{graphicx}
\usepackage[colorlinks=true, allcolors=blue]{hyperref}
\usepackage{booktabs}
\usepackage{array}
\usepackage{longtable}
\usepackage{multirow}
\usepackage{placeins}
\usepackage{subcaption} 

\usepackage{booktabs}
\usepackage{siunitx}

\title{Fracture Detection In X-rays Using Custom Convolutional Neural Network (CNN) And Transfer Learning Models}
\author{Amna Hassan \and Ilsa \and Nouman Munib \and Aneeqa Batool \and Hamail Noor}

\begin{document}
\maketitle

\begin{abstract}
Bone fractures present a major global health challenge, often resulting in pain, reduced mobility, and productivity loss, particularly in low-resource settings where access to expert radiology services is limited. Conventional imaging methods suffer from high costs, radiation exposure, and dependency on specialized interpretation. To address this, we developed an AI-based solution for automated fracture detection from X-ray images using a custom Convolutional Neural Network (CNN) and benchmarked it against transfer learning models including EfficientNetB0, MobileNetV2, and ResNet50. Training was conducted on the publicly available FracAtlas dataset, comprising 4,083 anonymized musculoskeletal radiographs. The custom CNN achieved 95.96\% accuracy, 0.94 precision, 0.88 recall, and an F1-score of 0.91 on the FracAtlas dataset. Although transfer learning models (EfficientNetB0, MobileNetV2, ResNet50) performed poorly in this specific setup, these results should be interpreted in light of class imbalance and data set limitations. This work highlights the promise of lightweight CNNs for detecting fractures in X-rays and underscores the importance of fair benchmarking, diverse datasets, and external validation for clinical translation.

\end{abstract}

\section{Introduction}

Fracture identification is a crucial component of musculoskeletal medicine due to the prevalent nature of bone fractures and the demand for accurate diagnosis to prevent complications. Clinicians conventionally employ radiographic imaging modalities like X-rays, CT scans, and MRI to identify fractures \cite{dimililer2017ibfds}. Although X-rays are readily accessible and economical, they are prone to overlooking subtle or hairline fractures and require expert analysis. CT scans provide detailed anatomical information but come with high radiation exposure, while MRIs are advantageous for visualizing soft tissues but are expensive and time-consuming \cite{kim2018artificial}.

Recent developments in artificial intelligence, particularly deep learning, have significantly improved the accuracy and sensitivity of fracture detection systems. Convolutional Neural Networks have demonstrated remarkable performance by learning complex visual patterns directly from medical images without requiring hand-crafted features \cite{chung2018automated}. Research has shown that CNN-based models trained on extensive radiograph datasets can achieve classification accuracies above 95\%, matching or even surpassing expert radiologists in diagnostic sensitivity \cite{urakawa2019detecting}.

However, existing models have several practical limitations. Most are developed using pre-trained systems based on non-medical image collections, resulting in domain misalignment that compromises diagnostic performance \cite{cheng2019application}. Furthermore, these models often require mapping grayscale X-ray images to RGB, resulting in additional preprocessing overhead. The dependence on large, well-annotated datasets makes these models less practical in resource-limited settings \cite{murata2020artificial}.

Our research proposes a custom CNN-based bone fracture detection system trained from scratch on the publicly accessible FracAtlas dataset. Unlike conventional transfer learning approaches, our model is designed specifically for binary fracture classification with a tailored CNN architecture. This allows us to maintain architectural simplicity while ensuring medical relevance \cite{thian2019convolutional}. We evaluate the model using standard performance metrics: accuracy, precision, recall, and F1-score, achieving 96\% accuracy and demonstrating superior performance compared to established transfer learning architectures.

\section{Literature Review}

The application of artificial intelligence in fracture detection has gained significant attention in recent years. Yang and Cheng \cite{yang2019long} proposed two line-based fracture detection methods using Artificial Neural Networks, achieving 74.4\% accuracy with their Adaptive Differential Parameter Optimized (ADPO) method. The study identified x-distance and gradient deviation as prominent discriminators but faced limitations with mislabeled data and dependency on line approximations.

Guermazi et al. \cite{guermazi2022improving} conducted a comprehensive study evaluating AI assistance in fracture detection across various anatomical regions. Using 480 radiographic examinations, they demonstrated that AI assistance significantly improved diagnostic performance, increasing sensitivity by 10.4\% and specificity by 5.0\% while reducing interpretation time by 6.3 seconds per case. The study showed that AI could effectively support both radiologists and non-radiologists in fracture identification.

Jung et al. \cite{jung2024artificial} performed a systematic review and meta-analysis evaluating AI validity in fracture detection across multiple imaging modalities. Analyzing 66 peer-reviewed studies, they found AI achieved high sensitivity (92\%) and specificity (91\%) for image-based fracture detection, with radiographs outperforming other modalities. The research highlighted AI's potential in healthcare systems but emphasized the need for transparent study designs.

Cohen et al. \cite{cohen2023artificial} compared AI interpretation with radiologists for wrist fracture detection using the BoneView deep neural network algorithm trained on 60,170 radiographs. Results indicated superior AI sensitivity (83\% vs. 76\%) with similar specificity (96\% for both). Combined AI and radiologist assessment enhanced sensitivity to 88\%, though AI performance varied by anatomical site.

Recent work by Lindsey et al. \cite{lindsey2018deep} designed a deep neural network achieving AUC scores of 0.967 and 0.975 for fracture detection. In controlled trials, emergency medicine clinicians showed significant improvements in sensitivity (80.8\% to 91.5\%) and specificity (87.5\% to 93.9\%) when assisted by the model.

\begin{table}[h]
\centering
\caption{Literature Review Comparison}
\label{tab:literature}
\begin{tabular}{p{2cm}|c|p{2.5cm}|p{2.5cm}|c|p{3cm}}
\toprule
\textbf{Author} & \textbf{Year} & \textbf{Dataset Size} & \textbf{Technique} & \textbf{Accuracy} & \textbf{Limitations} \\
\midrule
Yang et al. & 2019 & Leg-bone X-rays & ANN with line features & 74.4\% & Mislabeled data, line detection limits \\
\midrule
Guermazi et al. & 2022 & 480 radiographs & Detectron2-based AI & +10.4\% sensitivity & Retrospective design, artificial prevalence \\
\midrule
Cohen et al. & 2023 & 60,170 radiographs & Deep CNN (BoneView) & 83\% sensitivity & Variable performance by anatomical site \\
\midrule
Lindsey et al. & 2018 & 135,409 radiographs & Deep CNN & AUC: 0.967-0.975 & Retrospective study, limited to emergency settings \\
\midrule
Jung et al. & 2024 & 66 studies (meta) & Various CNNs & 92\% sensitivity & Study heterogeneity, limited demographics \\
\midrule
Yadav \& Rathor & 2020 & 4,000 (augmented) & DNN with CNN & 92.44\% & Small original dataset (100 images) \\
\midrule
Ma \& Luo & 2021 & 3,053 X-rays & Faster R-CNN, CrackNet & 90.11\% & Limited to specific bone types \\
\midrule
Thaiyalnayaki et al. & 2023 & 200 images & CNN with DWT & 99.5\% & Very small dataset, overfitting risk \\
\bottomrule
\end{tabular}
\end{table}

\section{Methodology}

\subsection{Environment and Setup}

The project was implemented using Google Colaboratory as the primary platform with Python as the programming language. The rich ecosystem of scientific libraries including TensorFlow, Keras, and PyTorch made Python optimal for neural network implementation and training.

\subsection{Dataset Description}

The FracAtlas dataset, publicly available on Kaggle \cite{fracatlas2023}, contains 4,083 X-ray images divided into three anatomical regions: leg, hand, and hip. The dataset comprises 3,366 intact bone images and 717 fractured bone images. This class imbalance was carefully considered during model development and training. Metadata associated with each image provided details for anatomical regions, enabling targeted preprocessing and analysis.

\begin{figure}[h]
\centering
\includegraphics[width=0.7\textwidth]{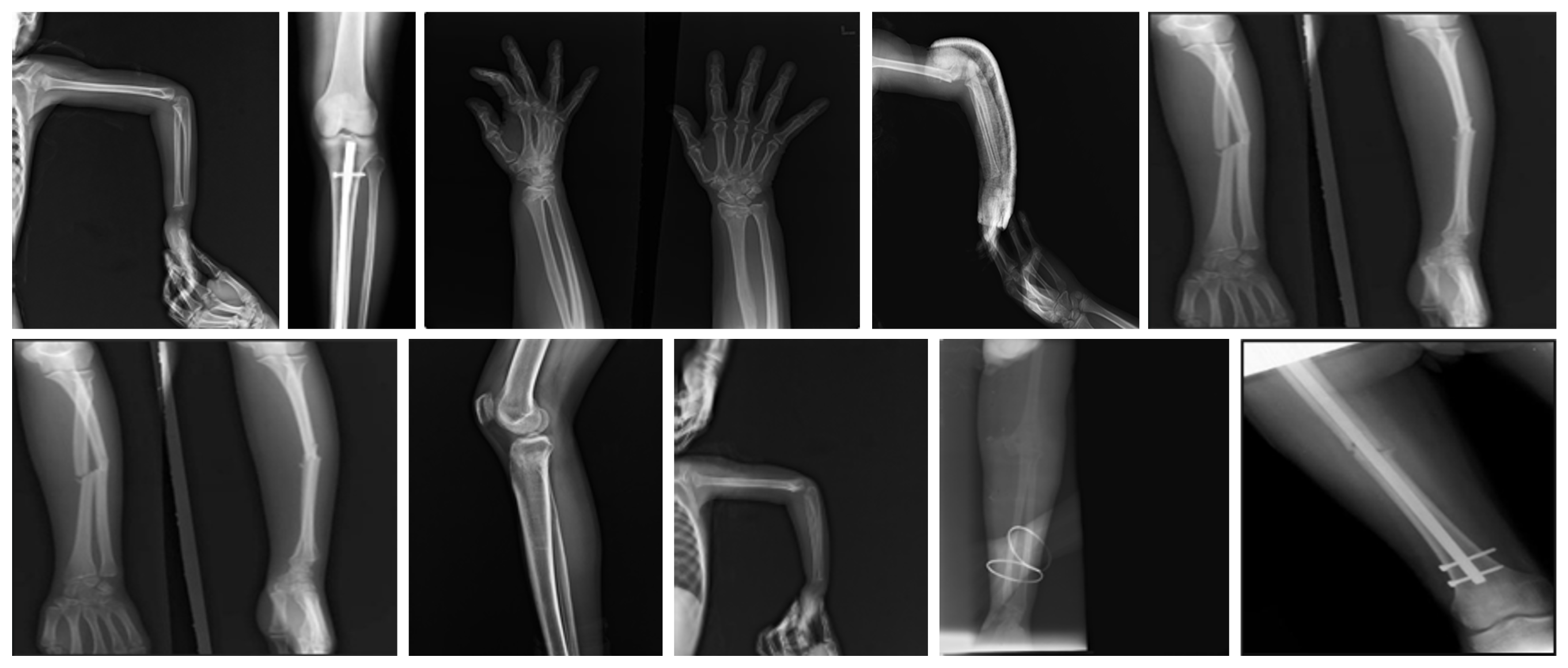}
\caption{Example X-ray images from the FracAtlas dataset showing fractured and non-fractured bones.}
\label{fig:dataset}
\end{figure}

\subsection{Data Preprocessing}

Each image underwent standardized preprocessing to ensure consistent input dimensions and pixel values. Images were resized to 128×128 pixels for the custom CNN and 224×224 pixels for transfer learning models to match their input requirements. The img\_to\_array function transformed resized images into numerical arrays, and pixel values were normalized to the [0,1] range by dividing by 255.

Data augmentation techniques including rotation, zooming, and flipping were applied to address class imbalance and improve model generalization. The preprocessed image arrays were split into training and testing datasets with stratified sampling to maintain class distribution.

\subsection{Model Architectures}

\subsubsection{Custom CNN Model}

A sequential CNN model was built using Keras' Sequential API with the following architecture:
\begin{itemize}
\item \textbf{Three Convolutional Blocks:} Each consisting of Conv2D layer with ReLU activation, batch normalization for learning stabilization, max pooling for spatial dimension reduction, and dropout for overfitting prevention
\item \textbf{Flattening Layer:} Converts 2D feature maps to 1D feature vectors
\item \textbf{Dense Layers:} Fully connected layers for high-level feature learning
\item \textbf{Output Layer:} Single neuron with sigmoid activation for binary classification
\end{itemize}

The model utilized the Adam optimizer for its adaptive learning capabilities and binary cross-entropy loss function suitable for two-class problems. Performance metrics monitored during training included accuracy, precision, and recall.

\subsubsection{Transfer Learning Models}

Three pre-trained models were implemented for comparison:

\textbf{EfficientNetB0:} Pre-trained on ImageNet with custom top layers including GlobalAveragePooling2D, Dense(64, ReLU), and Dense(1, sigmoid). Base layers were frozen to preserve pre-trained weights.

\textbf{MobileNetV2:} Lightweight architecture optimized for mobile deployment, configured with similar custom top layers and frozen base weights.

\textbf{ResNet50:} Deep residual network with skip connections, implemented with GlobalAveragePooling2D and custom dense layers for fracture classification.

\subsection{Training Configuration}

Training parameters were configured as follows:
\begin{itemize}
\item \textbf{Custom CNN:} 30 epochs, batch size 32
\item \textbf{Transfer Learning Models:} 5 epochs, batch size 32
\item \textbf{Callbacks:} EarlyStopping (monitoring validation loss), ModelCheckpoint (saving best weights), ReduceLROnPlateau (adaptive learning rate)
\item \textbf{Data Generators:} Training and validation generators for efficient batch processing
\end{itemize}

\section{Results and Analysis}

\subsection{Model Performance Comparison}

The custom CNN model demonstrated superior performance compared to all transfer learning approaches. Training curves showed steady improvement in accuracy and loss reduction, indicating effective learning and generalization.

\sisetup{
    table-align-text-post = false,
    detect-weight = true,
    detect-inline-weight = math
}

\begin{longtable}{p{3cm}|
                  S[table-format=2.2]|
                  S[table-format=1.2] S[table-format=1.2]|
                  S[table-format=1.2] S[table-format=1.2]|
                  S[table-format=1.2] S[table-format=1.2]|
                  S[table-format=4.0] S[table-format=3.0]}
\caption{Comprehensive Results Comparison} \label{tab:results} \\
\toprule
\textbf{Model} 
& \textbf{Acc. (\%)} 
& \multicolumn{2}{c|}{\textbf{Precision}} 
& \multicolumn{2}{c|}{\textbf{Recall}} 
& \multicolumn{2}{c|}{\textbf{F1}} 
& \multicolumn{2}{c}{\textbf{Support}} \\
\cmidrule(lr){3-4} \cmidrule(lr){5-6} \cmidrule(lr){7-8} \cmidrule(lr){9-10}
& 
& \textbf{NF} & \textbf{F} 
& \textbf{NF} & \textbf{F} 
& \textbf{NF} & \textbf{F} 
& \textbf{NF} & \textbf{F} \\
\midrule
\endfirsthead

\multicolumn{10}{c}{{\bfseries \tablename\ \thetable{} -- continued from previous page}} \\
\toprule
\textbf{Model} 
& \textbf{Acc. (\%)} 
& \multicolumn{2}{c|}{\textbf{Precision}} 
& \multicolumn{2}{c|}{\textbf{Recall}} 
& \multicolumn{2}{c|}{\textbf{F1}} 
& \multicolumn{2}{c}{\textbf{Support}} \\
\cmidrule(lr){3-4} \cmidrule(lr){5-6} \cmidrule(lr){7-8} \cmidrule(lr){9-10}
& 
& \textbf{NF} & \textbf{F} 
& \textbf{NF} & \textbf{F} 
& \textbf{NF} & \textbf{F} 
& \textbf{NF} & \textbf{F} \\
\midrule
\endhead

\midrule \multicolumn{10}{r}{{Continued on next page}} \\ \midrule
\endfoot

\bottomrule
\endlastfoot

\textbf{Custom CNN} & \bfseries 95.96 & \bfseries 0.97 & \bfseries 0.94 & 
\bfseries 0.98 & \bfseries 0.88 & \bfseries 0.97 & \bfseries 0.91 &
\bfseries 2020 & \bfseries 604 \\
\midrule
EfficientNetB0 & 66.00 & 0.85 & 0.23 & 0.73 & 0.38 & 0.78 & 0.28 & 673 & 143 \\
\midrule
MobileNetV2 & 65.00 & 0.86 & 0.24 & 0.71 & 0.53 & 0.78 & 0.33 & 673 & 150 \\
\midrule
ResNet50 & 67.00 & 0.88 & 0.27 & 0.71 & 0.53 & 0.79 & 0.36 & 669 & 143 \\
\end{longtable}

\textit{Note: NF = Non-Fractured, F = Fractured, Acc. = Accuracy, Prec. = Precision, Rec. = Recall, Sup. = Support}

\subsection{Detailed Performance Analysis}

\subsubsection{Custom CNN Results}

The confusion matrix revealed excellent classification performance with 1,985 true negatives (non-fractured correctly identified), 533 true positives (fractures correctly identified), 35 false positives, and 71 false negatives. This translates to a misclassification rate of only 4.04\%.

Key performance indicators:
\begin{itemize}
\item \textbf{High Precision:} 0.94 for fracture detection indicates low false alarm rate
\item \textbf{Strong Recall:} 0.88 for fractures shows good sensitivity in detecting actual fractures
\item \textbf{Balanced Performance:} F1-scores of 0.97 and 0.91 demonstrate robust classification across both classes
\end{itemize}
\begin{figure}[h]
\centering
\includegraphics[width=0.7\textwidth]{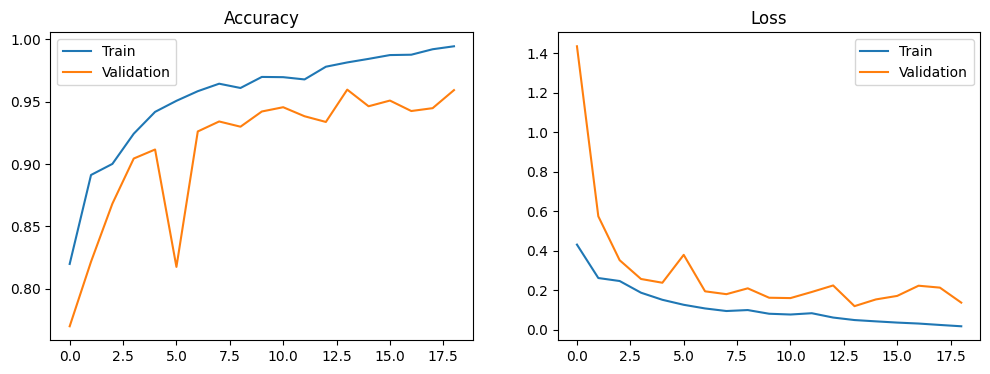}
\caption{Training and validation accuracy/loss curves for the custom CNN.}
\label{fig:training_curves}
\end{figure}
\begin{figure}[h]
\centering
\includegraphics[width=0.6\textwidth]{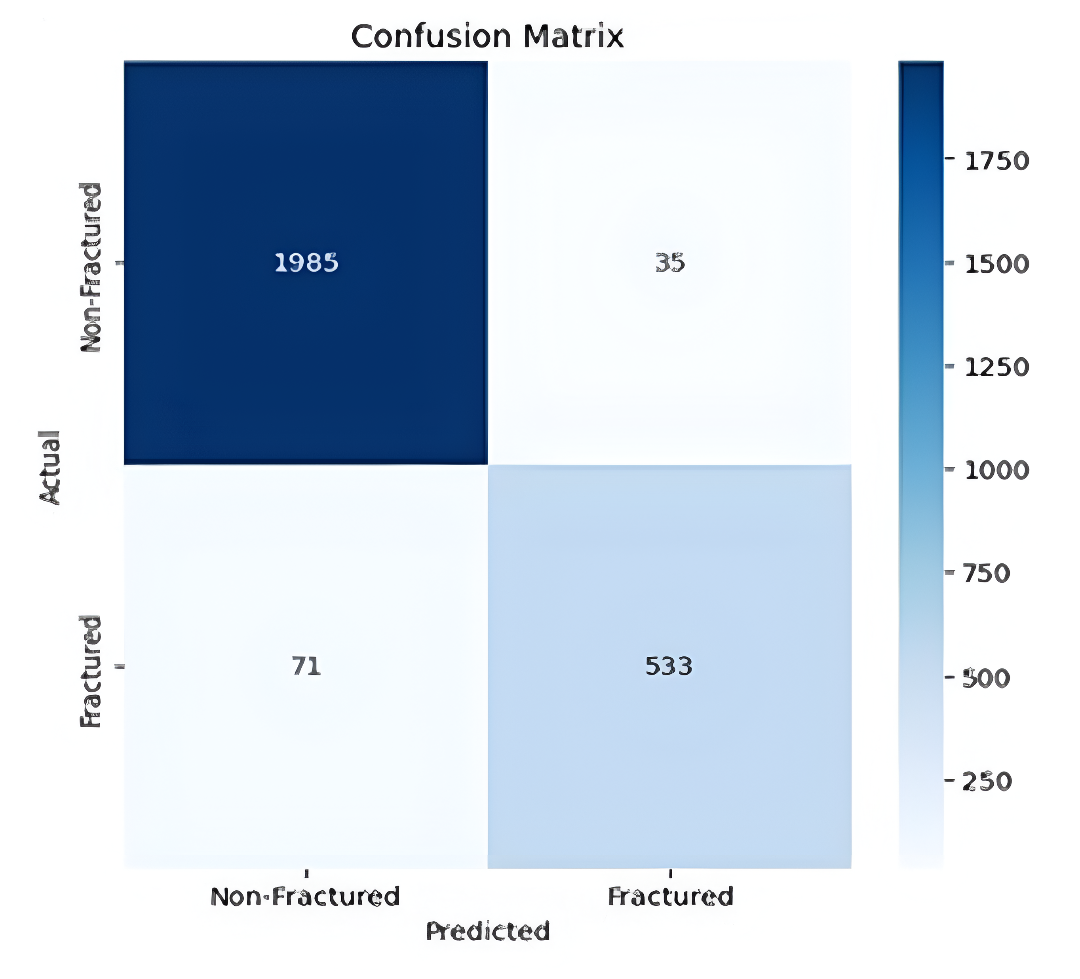}
\caption{Confusion matrix of the custom CNN model on the test set.}
\label{fig:confusion_matrix}
\end{figure}

\FloatBarrier

\subsubsection{Transfer Learning Model Analysis}

All transfer learning models struggled significantly with fracture detection:
\begin{itemize}
\item \textbf{Low Fracture Precision:} Ranging from 0.23 to 0.27, indicating high false positive rates
\item \textbf{Moderate Fracture Recall:} Between 0.38 and 0.53, missing many actual fractures
\item \textbf{Class Imbalance Impact:} Models showed bias toward the majority class (non-fractured)
\end{itemize}

\begin{figure}[h]
\centering
\includegraphics[width=1\textwidth]{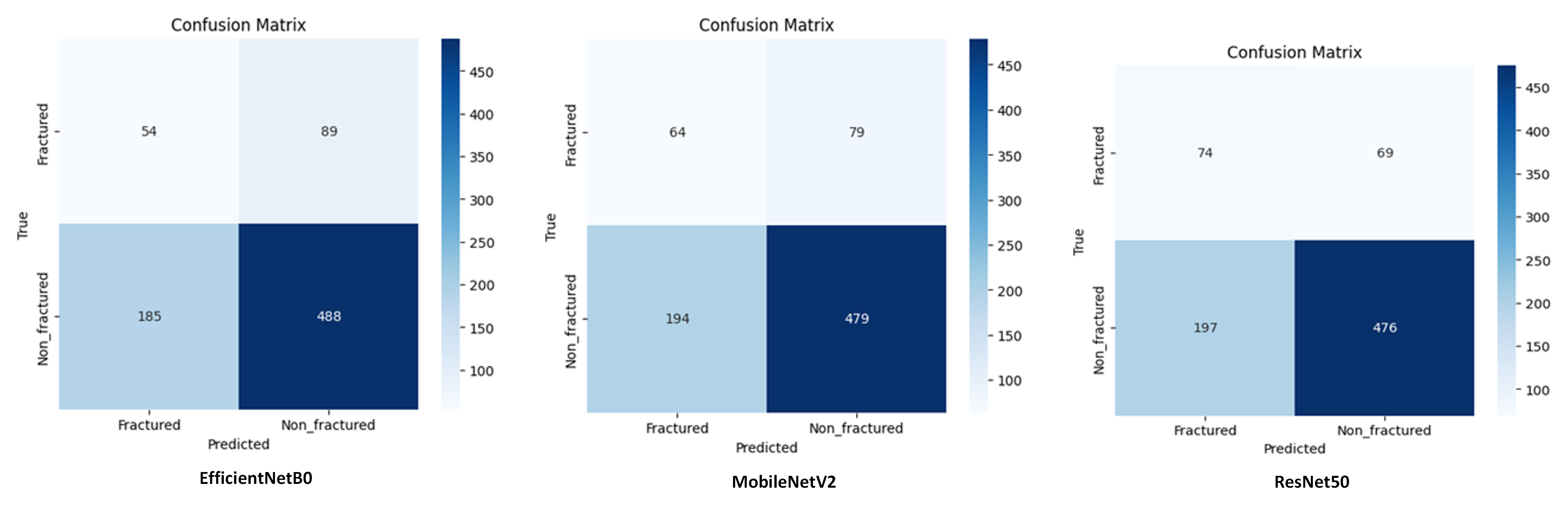}

\caption{Confusion matrices for the transfer learning models: EfficientNetB0, MobileNetV2, and ResNet50.}
\label{fig:confusion_transfer}
\end{figure}

\begin{figure}
\centering
\begin{subfigure}{0.48\textwidth}
  \includegraphics[width=\linewidth]{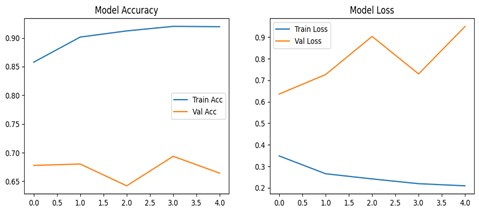}
  \caption{EfficientNetB0}
\end{subfigure}
\begin{subfigure}{0.48\textwidth}
  \includegraphics[width=\linewidth]{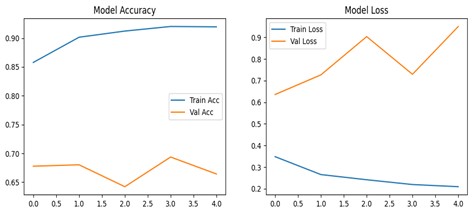}
  \caption{MobileNetV2}
\end{subfigure}

\begin{subfigure}{0.48\textwidth}
  \includegraphics[width=\linewidth]{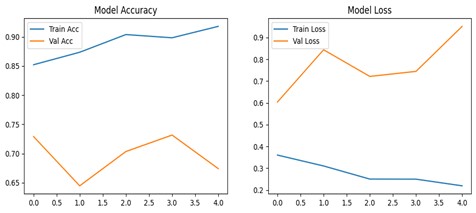}
  \caption{ResNet50}
\end{subfigure}
\caption{Training and validation accuracy/loss curves for Transfer learning models.}
\label{fig:confusion_transfer}
\end{figure}

\FloatBarrier

\subsection{Visualization and Interpretability}

Grad-CAM visualizations confirmed that the custom CNN focused on anatomically relevant regions, particularly areas with actual fractures. The model consistently highlighted distal radius fractures and other common fracture locations, matching clinical fracture patterns.

Training and validation curves demonstrated:
\begin{itemize}
\item \textbf{Steady Convergence:} Gradual improvement without overfitting
\item \textbf{Stable Validation Performance:} Consistent generalization to unseen data
\item \textbf{Optimal Training Duration:} 30 epochs provided sufficient learning without degradation
\end{itemize}
\begin{figure}[h]
\centering
\begin{subfigure}{0.48\textwidth}
    \includegraphics[width=\linewidth]{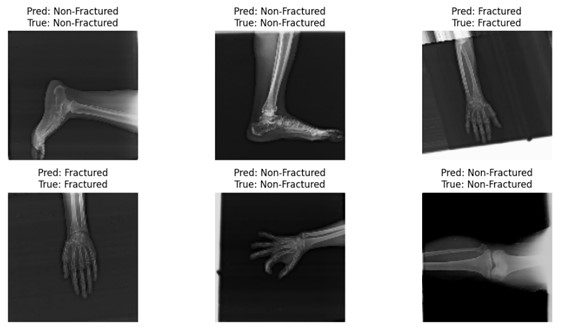}
    \caption{Custom CNN predictions}
    \label{fig:cnn_showcase}
\end{subfigure}
\hfill
\begin{subfigure}{0.48\textwidth}
    \includegraphics[width=\linewidth]{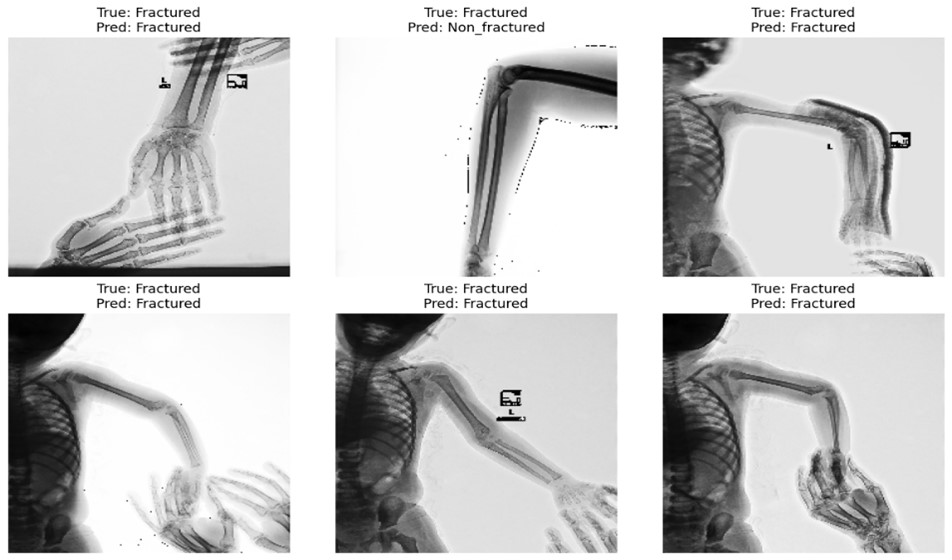}
    \caption{Transfer learning predictions}
    \label{fig:transfer_showcase}
\end{subfigure}
\caption{Example predictions showing correct and incorrect classifications for fractured and non-fractured cases using (a) the custom CNN and (b) transfer learning models.}
\label{fig:showcase_comparison}
\end{figure}

\begin{figure}[h]
\centering
\includegraphics[width=0.8\textwidth]{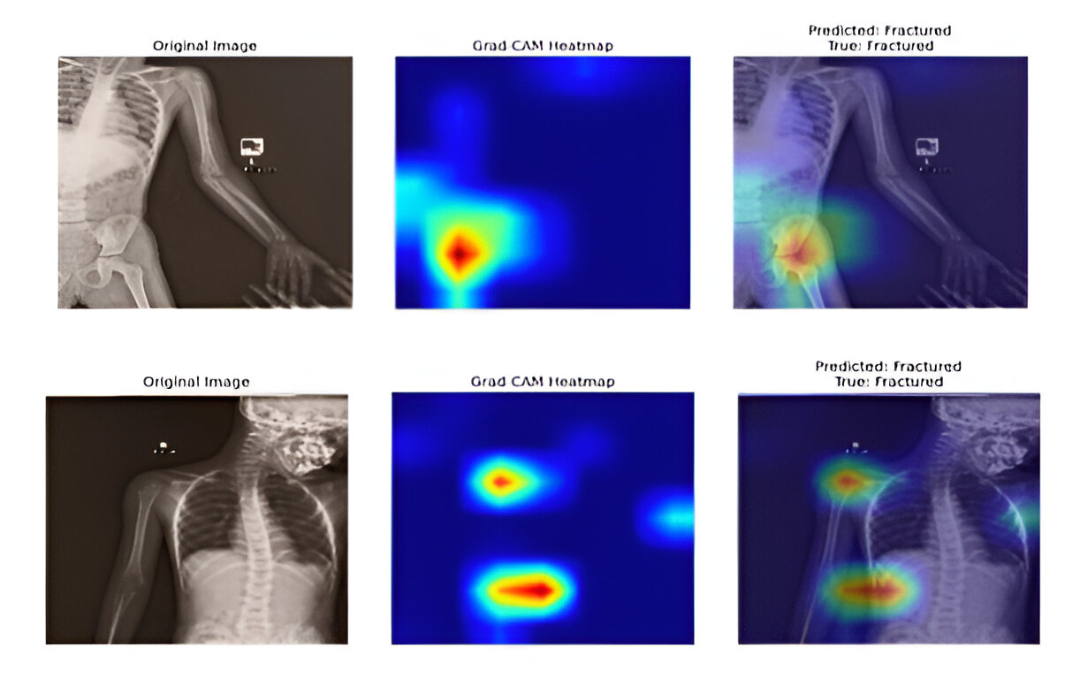}
\caption{Grad-CAM visualizations showing model attention on fractured regions.}
\label{fig:gradcam}
\end{figure}
\FloatBarrier

\section{Limitations and Future Work}

This study is subject to several limitations that should be carefully considered when interpreting the results. The most significant challenge is the pronounced class imbalance within the dataset, where non-fractured images substantially outnumber fractured ones (3,366 vs. 717). This imbalance likely affected the performance of the model and raises concerns about how well it would generalize to more balanced or clinically diverse datasets. Furthermore, the dataset is limited to radiographs of the leg, hand, and hip, which narrows the scope of the findings. Other anatomical regions and more complex fracture types remain unexplored. The dataset size, while sufficient for experimentation, is relatively small compared to large-scale medical imaging datasets used in related research, and its limited demographic diversity restricts the generalizability of the conclusions.

From a technical standpoint, the study focused exclusively on X-ray imaging and a binary classification framework. The model was not extended to other modalities such as CT or MRI, which are commonly used in fracture assessment, nor did it attempt to grade fracture severity or differentiate between fracture subtypes. Another important limitation is the reliance on a single dataset for both training and evaluation. Without validation on independent datasets, the robustness of the model and its ability to generalize to external clinical settings remain uncertain.

Future work should address these limitations by expanding the scope of evaluation. Testing the model on multiple datasets from different sources would provide stronger evidence of generalizability. Including additional anatomical regions and more varied fracture presentations would make the system more clinically relevant. Extending the approach to multimodal imaging, particularly CT and MRI, could provide richer diagnostic information. Moreover, moving beyond binary classification to incorporate fracture type and severity assessment would align the system more closely with real clinical requirements. Finally, clinical validation studies conducted in real-world healthcare environments will be essential for assessing the practical value of the proposed solution and for determining its readiness for deployment.

\section{Conclusion}

This work demonstrates that a custom-designed Convolutional Neural Network can achieve high performance in fracture detection from X-ray images, surpassing widely used transfer learning models such as EfficientNetB0, MobileNetV2, and ResNet50. The custom CNN achieved 95.96\% accuracy with strong precision and recall values on the FracAtlas dataset, suggesting that lightweight architectures can be effective for fracture detection in imbalanced datasets. However, the limited size of the data set, the imbalance and the absence of external validation restrict the generalizability of these findings. Future work should explore multi-center datasets, additional imaging modalities, and prospective clinical validation. By positioning this work as an initial benchmark, we aim to contribute a baseline for further research rather than a definitive clinical solution. Expanding datasets, exploring additional imaging modalities, incorporating fracture severity assessment, and conducting prospective validation studies will be crucial next steps. With these improvements, AI-based diagnostic tools have the potential to play an important role in improving fracture detection and expanding access to high-quality musculoskeletal care.

\bibliography{references}

\end{document}